\documentclass[letterpaper]{article}
\usepackage{aaai}
\usepackage{times}
\usepackage{helvet}
\usepackage{courier}
\usepackage{graphicx}
\usepackage{listings}
\begin{document}
\title{The Path to Autonomous Learners}
\author{Hanna Abi Akl\\
Data ScienceTech Institute\\
hanna.abi-akl@dsti.institute\\
}
\maketitle
\begin{abstract}
\begin{quote}
In this paper, we present a new theoretical approach for enabling domain knowledge acquisition by intelligent systems. We introduce a hybrid model that starts with minimal input knowledge in the form of an upper ontology of concepts, stores and reasons over this knowledge through a knowledge graph database and learns new information through a Logic Neural Network. We study the behavior of this architecture when handling new data and show that the final system is capable of enriching its current knowledge as well as extending it to new domains.  
\end{quote}
\end{abstract}

\section{Introduction}
Artificial intelligence has taken strides in enabling machines to perform tasks at near human-level.
This poses a question on the nature of the relation between machines and knowledge, specifically how far off machines are from acquiring knowledge in an intelligent way. Today's intelligent systems do not show the same signs of learning as humans, and while our methods of learning are by no means optimal, they at least enable us to adapt to new domains. Machines on the other hand still struggle when introduced to a new environment or domain and remain limited in performing human tasks.

The field of natural language, for example,  has considerably benefited from advances in neural models such as Large Language Models (LLMs). While these models have been able to rival human performance on many NLP tasks \cite{https://doi.org/10.48550/arxiv.2208.10264,https://doi.org/10.48550/arxiv.2111.01243,https://doi.org/10.48550/arxiv.2207.14382}, they still present shortcomings, most notably in their inability to reason and their over-reliance on huge amounts of training data to learn.

These shortcomings in LLMs have led to questions over their long-term capabilities, most notably how much data they should be provided to generalize their learning to almost any domain, how much scaling is required to accommodate this increase in data and whether or not these factors are enough to exhibit some sort of domain-adaptive learning like humans \cite{https://doi.org/10.48550/arxiv.2206.10498}. In the face of these apparent problems neural models tend to suffer from, a tide of symbolic applications has resurfaced, and research combining both neural and symbolic (dubbed neuro-symbolic) models has emerged. These models aim to leverage the power of neural networks while imbuing them with a rule-based framework \cite{https://doi.org/10.48550/arxiv.2109.06133}. The main upsides are of course to enhance the learning capabilities of these models (and be able to trace the way they learn, something that is notoriously difficult to do with black-box language models) and enable them to perform while relying less on data, thereby reducing the massive data set requirement.

In the vein of neuro-symbolic research, Logic Neural Networks (LNNs) have emerged as the perfect candidates to reconcile both neural and symbolic learning approaches \cite{https://doi.org/10.48550/arxiv.2006.13155}. While harnessing the capabilities of neural networks, LNNs are also capable of reasoning based on first-order logical rules, which makes it inherently easy for them to understand and derive concepts such as equivalence, negation and implication.

In order to make sense of acquired knowledge, knowledge graphs have been shown to map information faithfully by modeling concepts as nodes and connections between them as edges \cite{https://doi.org/10.48550/arxiv.2110.08012,Ji_2022}. This makes them attractive tools to turn to for storing and linking data from different sources.

In this paper, we propose a system that will reshape the manner in which machines acquire and extrapolate knowledge. We model our approach by placing human learning at its center, i.e., we draw inspiration from human intelligence by breaking down human learning into 3 main components: minimal knowledge, reasoning and learning. Each component is treated in a section as part of our final solution. We demonstrate how machines can benefit from this approach by starting with minimum knowledge and presenting 2 use cases: reasoning to enrich an existing knowledge set and extending learning to new domains.

The rest of this paper is organized as follows. We discuss related work, introduce the technical details of our proposed approach, present practical use cases and finally provide the conclusions of our work.

\section{Related Work}

Graph Neural Networks (GNNs) have recently gained traction in artificial intelligence research. They have been shown to boost graph representation learning and achieve state-of-the-art performance on many human tasks like computer vision and language \cite{https://doi.org/10.48550/arxiv.2209.13232,Wu_2021}. Particularly, GNNs have been useful in explaining prediction performance for Deep Learning models and shifting away from the black-box architecture of learning systems \cite{https://doi.org/10.48550/arxiv.2207.12599}. 

The field of Neuro-symbolic Computing has also surged and paved the way for Neuro-Symbolic Artificial Intelligence \cite{https://doi.org/10.48550/arxiv.2210.15889}. This branch of AI emphasizes a knowledge-driven paradigm that promotes a strong generalization ability and interpretability \cite{https://doi.org/10.48550/arxiv.2111.08164}. Neuro-symbolic systems have also shown promising performances on many human tasks which may make them key to unlocking next-generation AI \cite{https://doi.org/10.48550/arxiv.2209.12618}. However, these models present their own pitfalls, most notably in their inability to handle unstructured data, their weak robustness which can render them difficult to scale and their slowness in reasoning which might raise performance issues depending on the applications.

Both GNNs and Neuro-symbolic systems seem to possess advantageous qualities that merit further exploration in the endeavor to make artificial models more intelligent. Recent studies point to a combined effort to reconcile both technologies to inherit the semantic representation of graph networks as well as the logical framework that powers Neuro-symbolic models \cite{https://doi.org/10.48550/arxiv.2003.00330}. This combination effectively makes for more interpretable models that operate on clear logical rules and are capable of symbolically representing information in an inter-operable network. This paper follows the logical progression of these hybrid models and builds on them.

\section{Proposed Approach}
Humans are intuitive and logical beings. They are capable of many intelligent functions such as retaining and storing information for later retrieval, finding relationships between different objects and explaining new ideas using mechanics like deduction or comparison. Our approach aims to align machine learning with human learning and hopes to emulate the features of the latter. For that, we model a pipeline that approximates the human stages of learning. We identify 3 principle stages:
\begin{itemize}
\item \textbf{Minimal Input Knowledge.} The basic amount of information humans start with. In our pipeline, this is the starting data containing the basic knowledge a machine should have. Here we base our approach on the hypothesis that every being starts with a set of information that constitutes the foundation for future learning, and we implement our system accordingly. The intuition behind this approach is that humans always seek to expand their knowledge, but no matter how much new information they retain this represents an infinitesimal amount of the existing knowledge in the world. By identifying and attributing a minimal knowledge set to our system, we aim to emulate the scenario of a human being that is continually learning.
\item \textbf{Reasoning.} This phase assumes the system already possesses some data or information that represents knowledge. At this stage, the system should be capable of reasoning, i.e., making sense of the existing data. This implies defining and differentiating data points (which we loosely call concepts), as well as correctly finding and tracing relations that bind them, i.e., reasoning over concepts that are linked together.
\item \textbf{Learning.} The system should be prepared to deal with changing reference frames. For that, it must be able to adapt to new domains. Adaptability is a hallmark of intelligence and a necessary ingredient for learning. A system that is capable of adapting to new types and sources of information is a system that showcases thinking and can therefore extend the scope of its knowledge. There should also be a feedback mechanism to incorporate every new piece of knowledge and update the existing information of the system.
\end{itemize}

\begin{figure*}[!ht]
    \centering
    \noindent{\includegraphics[width=\textwidth]{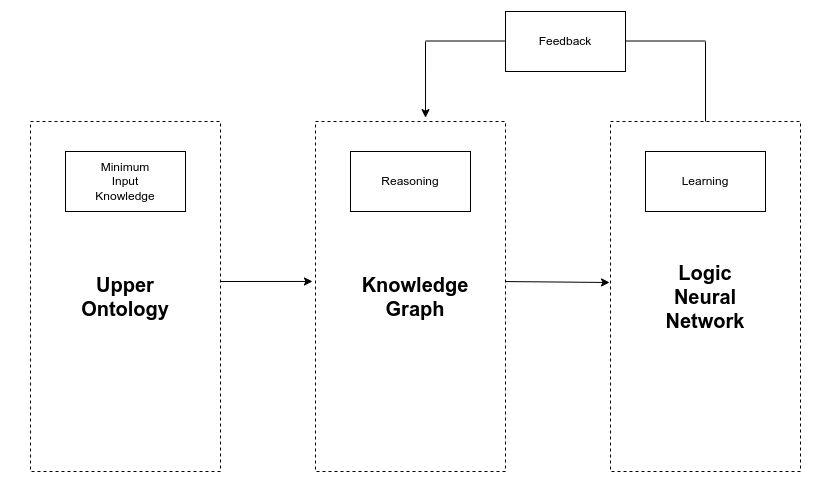}}
    \caption{Framework of our proposed approach}
\end{figure*}

Each principle stage is modeled by a component in our overall framework. The general architecture is shown in Figure 1. We elaborate each component in the following subsections.

\subsection{Minimal Input Knowledge}

In order to imbue our system with initial knowledge, we need to identify a basic set of representational knowledge that will serve as foundation for future learning. The criteria to consider in choosing this knowledge set are that it should be small (i.e., similarly to a human baby, the system need not start with huge amounts of data) and the data should be general enough to prevent specialization in a particular domain.

The problem of selecting or constructing the right knowledge set falls under the umbrella of \textit{Knowledge Engineering}. This problem has already been addressed at task-level, i.e., constructing the right knowledge set to perform well on a specific task \cite{https://doi.org/10.48550/arxiv.2202.01040}. Specifically, there is a branch of research in the field of knowledge engineering that tries to answer the question of deriving a general, consolidated knowledge set that represents the foundational knowledge of the universe. So far no such set has been derived, but several representative sets called \textit{Upper Ontologies} have been compiled for that purpose \cite{elmhadhbi:hal-02359706}.

An ontology is a formal representation of a set of information falling under the same theme or domain. An upper ontology is a high-level representation that transcends a specific domain and is general enough to cover multiple domains \cite{Mascardi2007ACO}. From the available upper ontologies, none constitutes the point of reference on its own, yet each of them possesses interesting features that makes it a reference depending on the case of study. For our system, we do not treat the problem of consolidating these ontologies, but rather focus our efforts on selecting the best one as a sufficient starting point for our Minimal Input Knowledge.

Our selection criteria are two-fold: size and availability. It is important that our system starts with a small information set like an early-stage human. Moreover, this knowledge set should be readily accessible to us. The choice falls on the Proton ontology specifically since it ticks both boxes: it is freely-available and small compared to the other upper ontologies. Proton consists of 25 classes and 77 properties and covers most of the high-level world concepts.

\subsection{Reasoning}

The purpose of the Reasoning module is to make sense of existing data. By representing the Proton ontology as a knowledge graph, we can map classes and properties to nodes and the relations linking them to edges. The advantage of using a knowledge graph is that it enforces a dynamic structure on the available information and allows easy navigation between concepts. In our architecture, nodes represent classes or high-level concepts, e.g., \textit{Person}. Edges represent relationships between classes, e.g., \textit{Person is a subclass of Agent}. Here the \textit{subclass of} relationship enables the system to extend the specific concept of \textit{Person} to the more general \textit{Agent} concept. This allows the system to classify new information by identifying the most likely class (or node) it might belong to.

Properties add a layer of precision to the representation of the knowledge set. They provide additional criteria to distinguish classes and can be help the system better define them. In the knowledge graph representation of the Proton ontology, properties are also represented as nodes and are linked to their relevant classes by edges.

The Reasoning module serves a dual purpose: retaining existing information and dynamically updating its network to accommodate new information. Whenever the system is exposed to a new concept, it will first treat this concept as an instance. We define an instance as a sub-node, i.e., a node derived from a class node. The system will first try to connect the instance somewhere along the existing graph network using the information it possesses on class nodes and their properties. If it fails to find a suitable class to link the instance to or has insufficient information about the instance, then it will treat it as a class by itself and add it as a concept node to the existing graph. The reasoning required to make these informed decisions relies on a logical framework, i.e., a set of rules, that is provided by the Learning module and is communicated to the Reasoning component through the Feedback module represented in Figure 1.

\subsection{Learning}

The Learning module is represented by a Logic Neural Network. The network instantiates a world model in which it learns the existing concepts and the connections between them through a set of first-order logic rules. These rules can be seen as the instruction set of the module to handle existing as well as incoming data and enables it to update the graph knowledge base by creating associations with new concepts.

For that, the module runs an inference on the set of rules (i.e., axioms) and the existing knowledge graph to derive meaning for the nodes and edges. Then, information can be queried in the form of predicates that either evaluate to \textit{True} or \textit{False} based on the rule set of the LLN. Using these mechanics, we can plan ahead for the system and anticipate its exposure to a new domain or environment by writing first-order logic rules that build on its Minimal Input Knowledge.

\subsection{Feedback}

The Feedback module is a wrapper code that establishes a connection with the graph knowledge base and provides an API to manipulate graph objects. The purpose of this component is to provide swift communication between the Reasoning and Learning modules by transforming logic rule inferences to graph database operations like query, insert and update on one end, and symbolic representations of new and existing data (i.e., variables in first-order logic) to graph objects, i.e., nodes and edges.

\section{Experiments}

To test the behavior of our framework, we devise 2 experiments designed to tackle different aspects of the learning process: enriching the current knowledge set with new information and integrating a new body of knowledge. We develop each experiment in the following subsections.

\subsection{Enriching knowledge set}

We define the Proton upper ontology as our starting knowledge set. We store the data in a knowledge graph such that classes and properties are represented by nodes. Class-class, class-property and property-property relations are represented by edges. In the Proton ontology, the only type of class-class relation is the \textit{subclass of} relationship. Class-property relations are represented differently: each property has a domain and a range, i.e., a mapping from one class to another. The domain references the source class and the range represents the target class of the property. Finally, for property-property relations, we choose to focus on the 2 most represented types: \textit{subproperty of} and \textit{inverse of}. Table 1 shows how class-subclass relationships are stored. In Table 2, we list all class-property relations. Tables 3 and 4 showcase the subPropertyOf and inverseOf relations between properties in our graph network.

\begin{table}[ht]
\begin{center}
\begin{tabular}{|l|l|l|l|l|l|}
\hline \bf Child & \bf Parent & \bf Child & \bf Parent  \\ 
\hline
Language & Abstract	& Event		& Entity \\
Language	& 	Entity &	JobPosition &	SocialPosition \\
Happening &		Entity &	JobPosition &		Situation \\
Organization &		Group &	JobPosition &		Happening \\
Organization 	 & Agent &	JobPosition &	Entity \\
Organization &		Object &	Group &		Agent \\
Organization &		Entity &	Group	& 	Object \\
ProductModel &		Object &	Group &		Entity \\
ProductModel &	Entity &	Document &		Inf.Res. \\
Location &		Object &	Document &		Statement \\
Location &	Entity &	Document &		Object \\
ContactInformation	 & 	Abstract &	Document	 & 	Entity \\
ContactInformation & 	Entity &	Abstract &		Entity \\
Inf.Res. &		Statement &	Role &		Situation \\
Inf.Res. &		Object &	Role &	Happening \\
Inf.Res. &		Entity &	Role & 	Entity \\
Service	& 	Object	 & Agent &		Object \\
Service &	Entity &	Agent &		Entity \\
Number &		Abstract &	Topic &		Abstract \\
Number &	Entity &	Topic &	 Entity \\
TimeInterval &		Happening &	Object &		Entity \\
TimeInterval &	 Entity &	Statement &		Object \\
Situation &		Happening &	Statement &		Entity \\
Situation &		Entity &	GeneralTerm &		Abstract \\
SocialPosition &		Situation &	GeneralTerm &		Entity \\
SocialPosition &		Happening &	Person &		Agent \\
SocialPosition &		Entity &	Person &		Object \\
Event &	Happening & Person &		Entity \\
\hline
\end{tabular}
\end{center}
\caption{Class subClassOf Relations}
\end{table}

\begin{table}[ht]
\begin{center}
\begin{tabular}{|l|l|l|l|l|l|}
\hline \bf Class & \bf Property & \bf Class & \bf Property \\ 
\hline
Happening	& 	End Time & Happening	& 	Pcpt in \\
Happening	&  	Ent. Pcptng  & Happening	& 	Start Time \\
Entity	& 	Located in  & Entity	& 	Name \\
Entity	& 	Involved in  & Entity	&  	Main Label \\
Entity	&  	Part of  & Entity	& 	Ent. Invld in \\
Entity	&	Description  & Org.	& 	Business as \\
Org.	& 	Etbld in & Org.	& 	Nb of Empl. \\
Org.	& 	Parent Org. of & Org.	& 	Etbld Date \\
Org.	& 	Subs. Org. of & Org.	& 	Rgstrd in \\
Pdct. Mdl.	& 	Produced by & Location	&  	Latitude \\
Location	& 	NIMA GNS Des. & Location	& 	Pop. Count \\
Location	& 	NIMA GNS UFI & Location	& 	Longitude \\
Location	& 	Subregion of & Inf.Res.	&	has Subject \\
Inf.Res.	& 	in Language & Inf.Res.	& 	Res. Format \\
Inf.Res.	& 	Dvd from Src & Inf.Res.	&  	Inf.Res. Cov. \\
Inf.Res.	& 	has Contributor & Inf.Res.	& 	Inf.Res. Rghts \\
Inf.Res.	& 	Inf.Res. Id & Inf.Res.	& 	has Date \\
Inf.Res.	& 	Title & Inf.Res.	& 	Res. Type \\
Service	& 	Operated by & Social Pos.	& 	Soc. Pos. Hldr \\
Job Pos.	& 	Holder & Job Pos.	& 	Held from \\
Job Pos.	& 	within Org. & Job Pos.	& 	Held to \\
Group	&  	has Member & Document	& 	Doc. Abstract \\
Document	& 	Dcmt Subttle & Role	& 	Role Holder \\
Role	& 	Role in & Agent	& 	Involved in \\
Agent	& 	is Legal Entity & Agent	& 	Part. Controls \\
Topic	& 	Subtopic of & Object	& 	is Owned by \\
Object	& 	Cnt. Info & Statement	& 	Valid from \\
Statement	& 	Valid until & Statement	& 	Stated by \\
Person	& 	is Boss of & Person	& 	has Relative \\
Person	&  	Soc. Pos. & Person	& 	Last Name \\
Person	& 	Given Name & Person	& 	has Pos. \\
Person	& 	First Name \\
\hline
\end{tabular}
\end{center}
\caption{Class-Property hasProperty Relations}
\end{table}

\begin{table}[ht]
\begin{center}
\begin{tabular}{|l|l|l|l|l|l|}
\hline \bf Source  & \bf Target & \bf Source  & \bf Target \\ 
\hline
Etbld in	& 	Located in & Doc. Abstct	& 	Desc. \\ 
Rgstrd in	& 	Located in &has Creator	& 	has Contr. \\ 
has Parent	& 	has Relative & Held to	& 	End Time \\ 
has Old Name	&	Name & has Siblg	& 	has Reltve \\ 
Invld in	& 	Ent. Invld in & Doc. Subttle	& Laconic Desc. \\ 
First Name	& 	Name & Title	&  	Name\\ 
has Employee	&  	has Member & Subregion of	& 	Part of \\ 
Held from	& 	Start Time & Doc. Author	& 	has Creator \\ 
Given Name	& 	Name & Last Name	& 	Name \\ 
has Spouse	& 	has Relative & Part. Owns	&  	Part. Controls \\ 
has Child	& 	has Relative & Owns	& 	Part. Owns \\ 
Subs. Org. of	& 	Part of & Laconic Desc.	& 	Desc. \\ 
Subregion of	& 	Located in & Pcpt in Happng	& 	Ent. Pcptng  \\
has Leader	& 	has Member & Parent Org. of	& 	Part. Controls \\ 
Doing Bsns as	& 	Name & Controls	& 	Part. Controls \\ 
\hline
\end{tabular}
\end{center}
\caption{Property subPropertyOf Relations}
\end{table}

\begin{table}[ht]
\begin{center}
\begin{tabular}{|l|l|}
\hline \bf Source & \bf Target\\ 
\hline
Soc. Pos. Holder		& 	has Soc. Pos. \\
has Parent		& 	has Child \\
has Soc. Pos.		& 	Soc. Pos. Holder \\
has Position		& 	Holder \\
Parent Org. of		& 	Subs. Org. of \\
Pcpt in Happng		& 	Involved in \\
Ent. Pcptng		&	Entity Involved in \\
\hline
\end{tabular}
\end{center}
\caption{Property inverseOf Relations}
\end{table}

We also add the following axioms (i.e., first-order logic rules) to our system as the basic rule set to enable it to make sense of its existing data and learn new information in the context of its graph network:
\begin{itemize}
\item propagate-class-instance-to-superclass (Axiom 1): $\forall x\forall y\forall z(isinstanceOf(x,y) \land subClassOf(y,z) \Longrightarrow (isinstanceOf(x,z)))$
\item propagate-class-property-to-instance (Axiom 2): $\forall x\forall y\forall z(isinstanceOf(x,y) \land propertyOf(z,y) \Longrightarrow (propertyOf(z,x)))$
\item propagate-subproperty-to-class (Axiom 3): $\forall x\forall y\forall z(subPropertyOf(x,y) \land propertyOf(y,z) \Longrightarrow (propertyOf(x,z)))$
\item propagate-inverse-to-class (Axiom 4): $\forall x\forall y\forall z(inverseOf(x,y) \land propertyOf(y,z) \Longrightarrow (propertyOf(x,z)))$
\end{itemize}

Using these axioms, the model should be able to make clever deductions like follow a chain of subClassOf relations from parent to child to grandchild node and deduce that the grandchild is a subclass of its parent and grandparent, propagate class properties to all instances of that class, understand that a subproperty and an inverse property are tied to a property and trace it back to its relevant class. We test our model's understanding by introducing 2 examples:
\begin{itemize}
\item \textbf{english}: We define this information as an instance of the \textit{Language} class. Since \textit{Language} is a subclass of \textit{Abstract} and \textit{Abstract} is a subclass of \textit{Entity}, the model correctly deduces that \textbf{english} is a subclass of both \textit{Abstract} and \textit{Entity} (Axiom 1) and adds these relations to its network.
\item \textbf{paris}: We define this information as an instance of the \textit{Location} class. The model applies the same reasoning to deduce that \textbf{paris} is also a subclass of \textit{Object} and \textit{Entity} since \textit{Location} is a subclass of \textit{Object} and \textit{Object} is a subclass of \textit{Entity} (Axiom 1). Additionally, \textit{Location} has the following properties: "nima gns unique feature identifier", "longitude", "population count", "subregion of", "nima gns designator" and "latitude". The instance \textbf{paris} also inherits these properties (Axiom 2).
\end{itemize}

\subsection{Extending to a new domain}

To test the adaptability of our system, we introduce it to a new set of knowledge. We choose another top-level ontology, the Basic Formal Ontology (BFO), to imbue our framework with as much general knowledge as possible. Another possibility would have been to provide a more domain-specific ontology to try to specialize our system.

The BFO ontology is another freely-available resource that contains 34 categories and 8 relations. By integrating this ontology into the framework's knowledge graph, we aim to measure where and how BFO concepts intersect with Proton concepts. Since the BFO is designed to promote interoperability between domains, its categories consist mainly of general concepts much like the Proton ontology.

Figure 2 displays the hierarchical structure of the BFO. 

\begin{figure*}[!ht]
    \centering
    \noindent{\includegraphics[width=\textwidth]{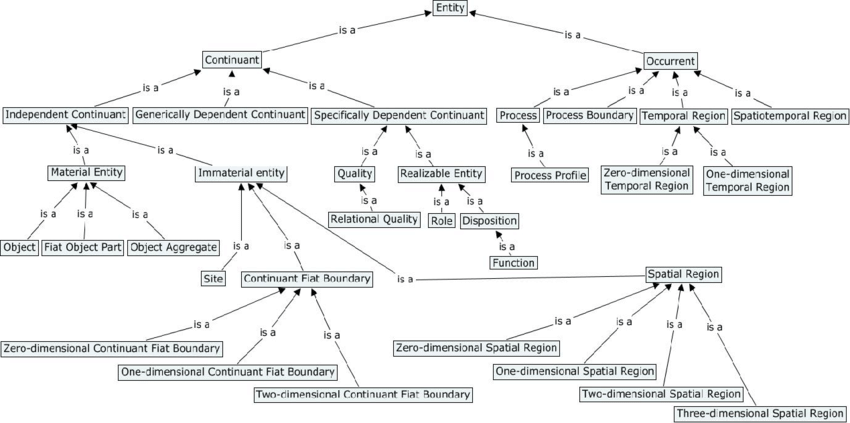}}
    \caption{Structure of the Basic Formal Ontology (BFO)}
\end{figure*}

We see that the BFO defines \textit{Entity} and \textit{Object} concepts, much like the Proton ontology. Our model should be able to identify these similarities and dynamically extend its network accordingly. By leaving the axioms in our system unchanged, we introduce a new \textbf{square} term and define it as an instance of \textit{Object}. But which \textit{Object} are we referring to? Is it the concept belonging to the Proton ontology or that of BFO? The answer is both. With the rule set at its disposal, our model should be capable of handling this ambiguity and deriving the connections and properties learned from both ontologies and attributing them to the \textbf{square} instance. The system produces the correct information and deduces that \textbf{square} is a subclass of \textit{Entity} from the Proton ontology and \textit{Material Entity} from the BFO (Axiom 1). It also attributes the "is owned by" and "has contact info" \textit{Object} properties to \textbf{square} (Axiom 2).

\subsection{Results}

From our experiments, we see that our system is capable of both reasoning and learning. The suggested framework shows that the model can reason since it can handle new incoming information and tie it to its data by using its graph network of classes and properties to augment its existing knowledge set.

Our architecture also proves the overall system can learn since it can incorporate additional domain information like new ontologies and integrate it to its knowledge base through a set of logical rules.

Since our experiments are performed exclusively with upper ontologies, we find that our proposed system also enables swift integration between them. This may lead to future work on the topic of creating a master top-level ontology that serves as a unique reference for all general knowledge.

Finally, to verify the sanity of our model's reasoning, we make the system output its learned network including the Proton ontology, the BFO and the examples we used for our experiments. The full model log results can be found in the Appendix section. We also publicly share the code containing the framework and experiments in this repository \footnote{https://github.com/HannaAbiAkl/AutonomousLearner}.

\subsection{Limitations}

Most data in the real world is unstructured, as opposed to formally-defined ontologies. One such form of unstructured data is natural language text. We can then ask the question of how well our system is equipped to handle natural language to extract relevant information from it or answer queries for example. If the ultimate objective of our proposed framework is to get closer to human intelligence, then these are some of the multitude of tasks it should be able to deal with.

However, this remains an open question that we haven't explored yet. Handling natural language requires at least an intermediate process to tokenize the words in the text or transform it into a logical set of information by means of relation extraction methods for example. There is also the question of the accuracy of such methods in retaining all the information from the original text. This is why this process is outside the scope of our research. Our question assumes such a functioning pipeline exists and asks how this information can and should be handled by our system.

Unfortunately, the strength of our framework might also be its greatest weakness. Since the model reasons in first-order logic, it expects predicative statements to be able to draw inferences from them. Transforming natural language text to first-order logic is an ongoing research \cite{inproceedings,chen2021neurallog}, but for now, this may well prove to be a limitation of our system. In case this transformation cannot happen, this weakness can be seen as a constraint rather than a liability in the sense that we will be required to formalize unstructured text before ingesting it in our framework.

Another approach would be to derive information from natural language using a grammar template. A problem with this method is that these templates should exist for all grammars and all languages. An example template can be found in the SUMO ontology, another upper-level ontology that is made available for us to use \cite{https://doi.org/10.48550/arxiv.2012.15835,https://doi.org/10.48550/arxiv.1808.04620}. The SUMO ontology is designed especially for research and applications in search, linguistics and reasoning. It is also mapped to the WordNet lexicon and is the largest public ontology in existence today with 13457 terms, 193812 axioms and 6055 rules. The English grammar template is a sub-graph of the ontology and consists of nodes and edges that can derive meaning from elements in text by linking them to semantic concepts. This implementation merits further exploration but our initial observations when trying to integrate the SUMO ontology in our framework is that it makes the model inference very slow due to its large size. Keeping our set of logical rules unchanged, ingestion of the Proton and BFO ontologies takes a few seconds compared to ingesting the SUMO ontology which takes several hours. This performance degradation presents itself as a limitation of our framework that raises optimization questions.

\section{Conclusion and Future Work}

In this paper, we propose a new theoretical approach for better machine learning. We draw inspiration from human intelligence and leverage the power of knowledge graphs and logic neural networks to create a hybrid framework capable of reasoning and learning with minimal input knowledge. We show that our system is capable of enriching its knowledge set by associating concept properties with new instances of its known classes. We also prove that the model is capable of extending its knowledge by integrating new domain information in its knowledge base and forming connections between related concepts via logical rule inference. These results, while early, deliver on the promise of adopting a neuro-symbolic approach in artificial intelligence and pave the way for future experiments to address interesting applications such as compiling a reference general ontology or understanding natural language more seamlessly. We hope this paper is a step toward creating autonomous learners truly capable of leveraging human-like intelligence.

\section{Appendix}

This section presents the full log results of the model's knowledge. The logs summarize the information retained by the system as well as the inferences drawn from the logic rules. They demonstrate how the model defines objects and their connections. Figure 3 shows all class-subclass relations by propagating the subClassOf relationship through the concept class hierarchy. Figure 4 shows all class-instance connections and class-property relationships. Figure 5 displays the model's inference on Axiom 4. Figure 6 displays the model's inference on Axioms 1 and 3. Figure 7 displays the model's inference on Axiom 2. Figure 8 showcases the subPropertyOf relation propagated throughout the model. Figures 9 and 10 shows the propertyOf relation between 2 nodes. Figures 11 and 12 log all Property nodes. Figures 13 and 14 show the subClassOf relation between 2 nodes. Figure 15 displays the instanceOf relationship between 2 nodes. Figures 16 and 17 show all existing Class and Instance nodes in the network.

\begin{figure*}[h]
    \centering
    \noindent{\includegraphics[width=\textwidth]{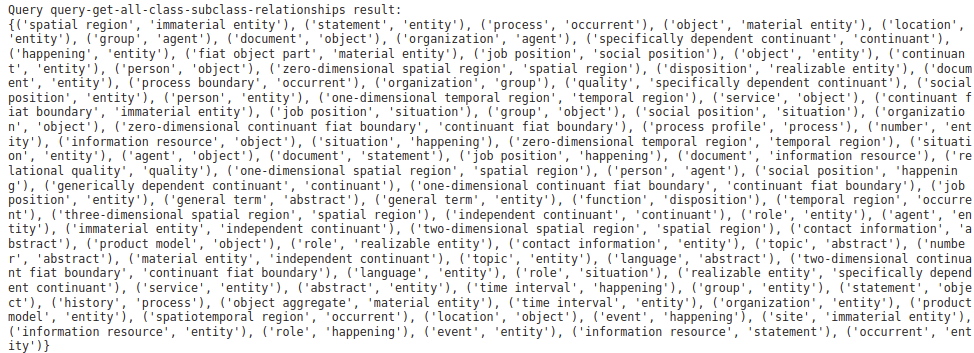}}
    \caption{Propagation of the subClassOf relation}
\end{figure*}

\begin{figure*}[!ht]
    \centering
    \noindent{\includegraphics[width=\textwidth]{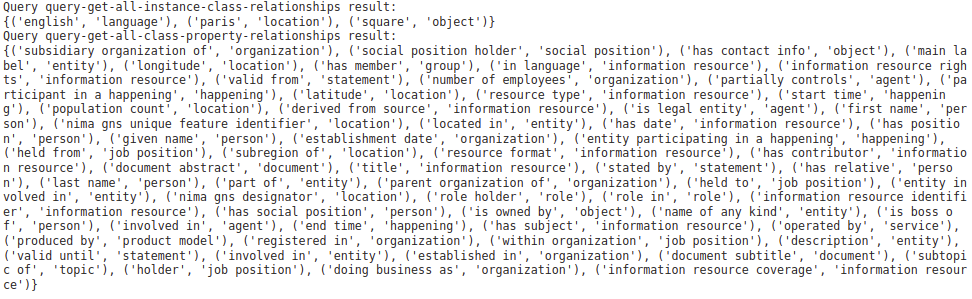}}
    \caption{Propagation of class instances and properties}
\end{figure*}

\begin{figure*}[!ht]
    \centering
    \noindent{\includegraphics[width=\textwidth]{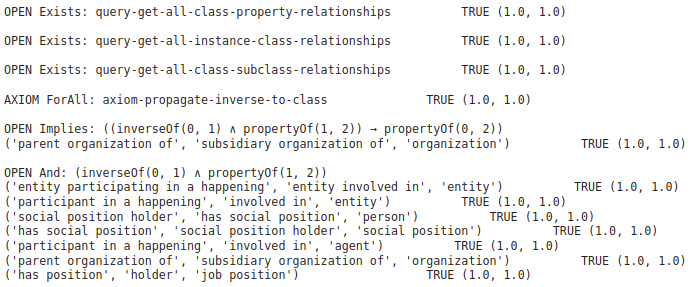}}
    \caption{Propagation of Axiom 4}
\end{figure*}

\begin{figure*}[!ht]
    \centering
    \noindent{\includegraphics[width=\textwidth]{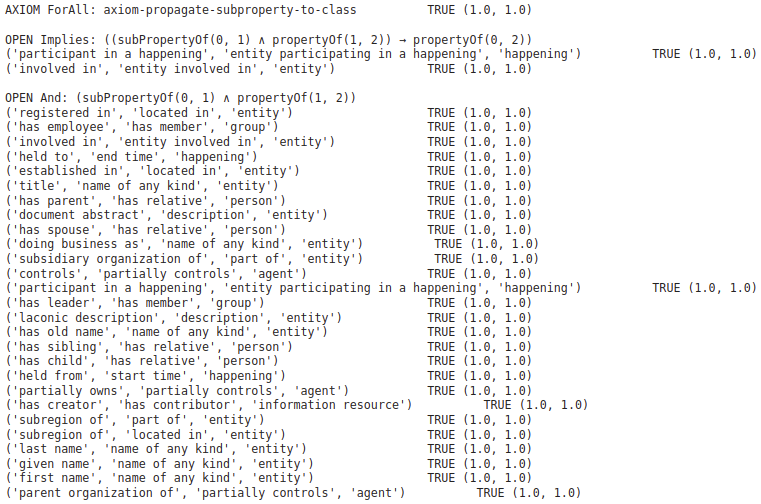}}
    \caption{Propagation of Axioms 1 and 3}
\end{figure*}

\begin{figure*}[!ht]
    \centering
    \noindent{\includegraphics[width=\textwidth]{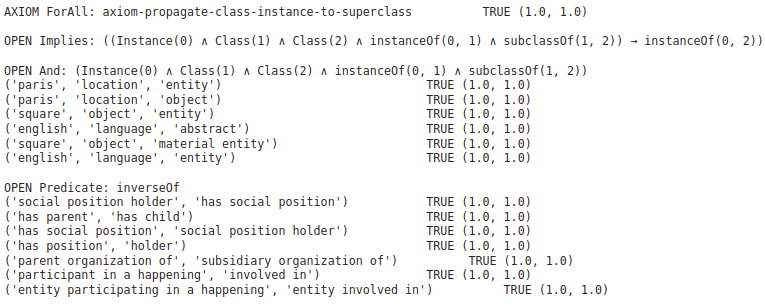}}
    \caption{Propagation of Axiom 2}
\end{figure*}

\begin{figure*}[!ht]
    \centering
    \noindent{\includegraphics[width=\textwidth]{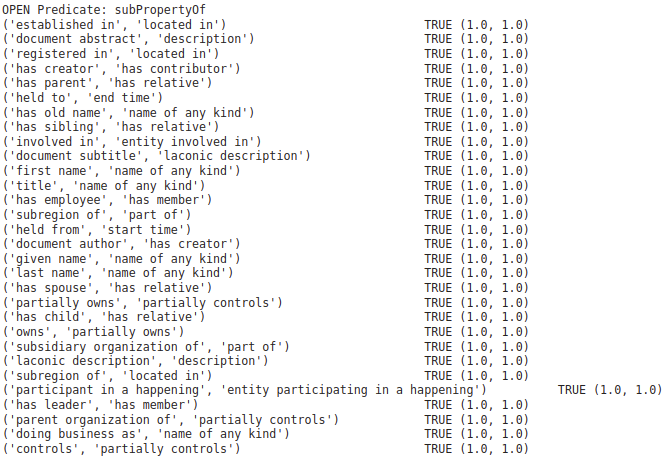}}
    \caption{Propagation of the subPropertyOf relation}
\end{figure*}

\begin{figure*}[!ht]
    \centering
    \noindent{\includegraphics[width=\textwidth]{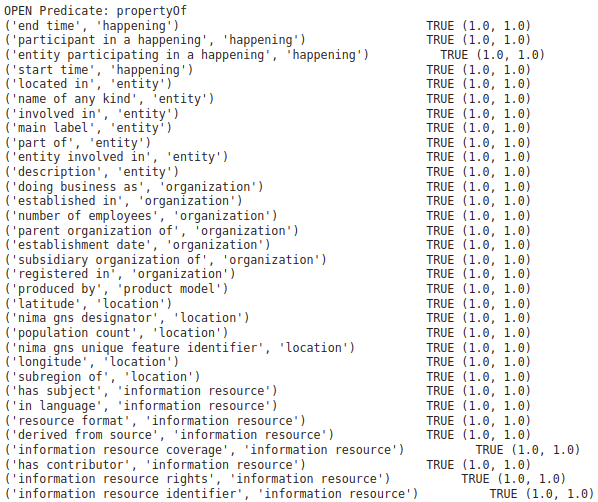}}
    \caption{Propagation of the propertyOf relation}
\end{figure*}

\begin{figure*}[!ht]
    \centering
\noindent{\includegraphics[width=\textwidth]{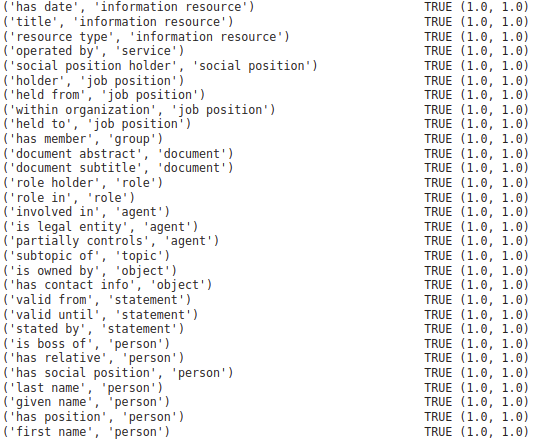}}
    \caption{Propagation of the propertyOf relation - continued}
\end{figure*}

\begin{figure*}[!ht]
    \centering
    \noindent{\includegraphics[width=\textwidth]{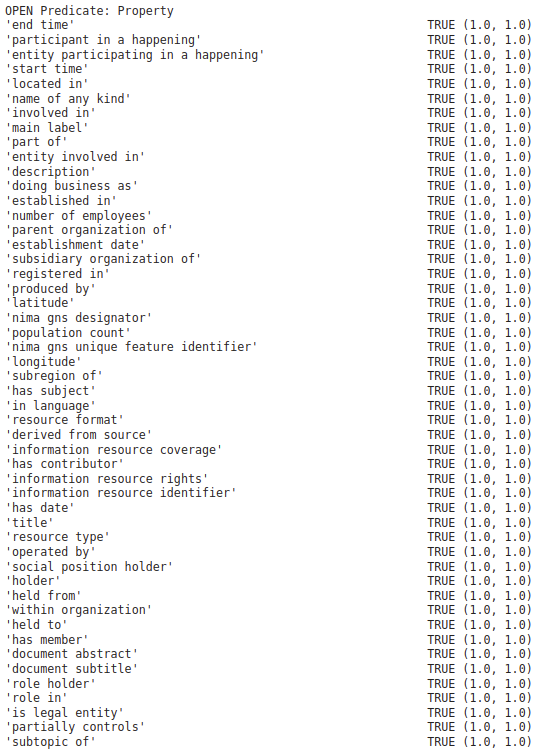}}
    \caption{Log of all Property nodes}
\end{figure*}

\begin{figure*}[!ht]
    \centering
    \noindent{\includegraphics[width=\textwidth]{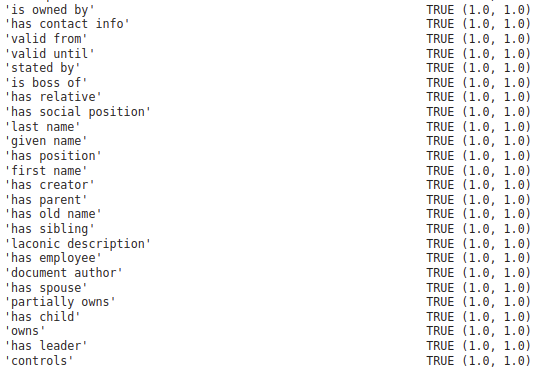}}
    \caption{Log of all Property nodes - continued}
\end{figure*}

\begin{figure*}[!ht]
    \centering
    \noindent{\includegraphics[width=\textwidth]{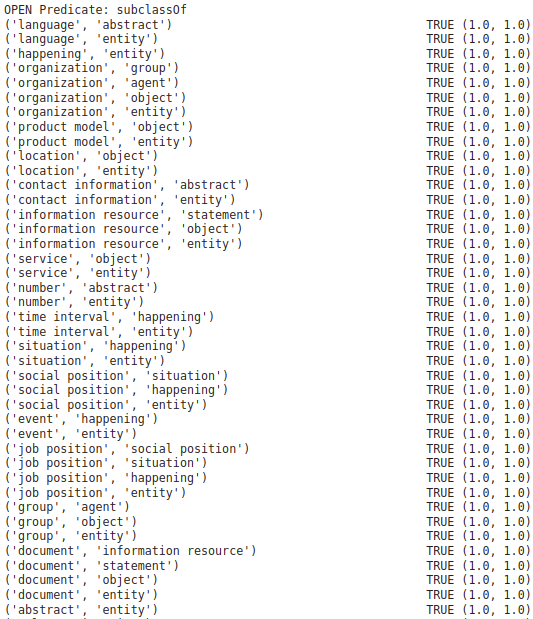}}
    \caption{Node-node subClassOf relations}
\end{figure*}

\begin{figure*}[!ht]
    \centering
    \noindent{\includegraphics[width=\textwidth]{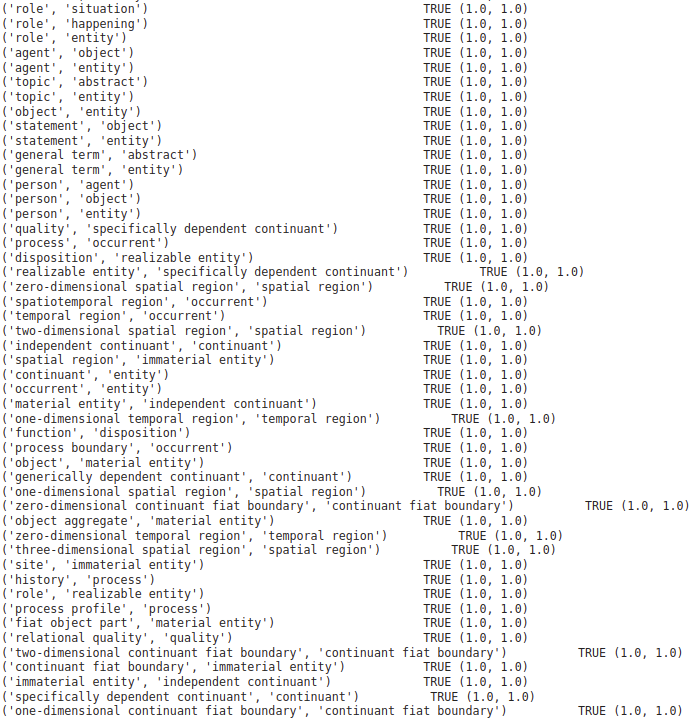}}
    \caption{Node-node subClassOf relations - continued}
\end{figure*}

\begin{figure*}[!ht]
    \centering
    \noindent{\includegraphics[width=\textwidth]{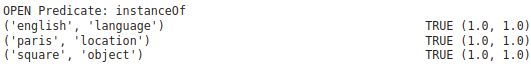}}
    \caption{Node-node instanceOf relations}
\end{figure*}

\begin{figure*}[!ht]
    \centering
    \noindent{\includegraphics[width=\textwidth]{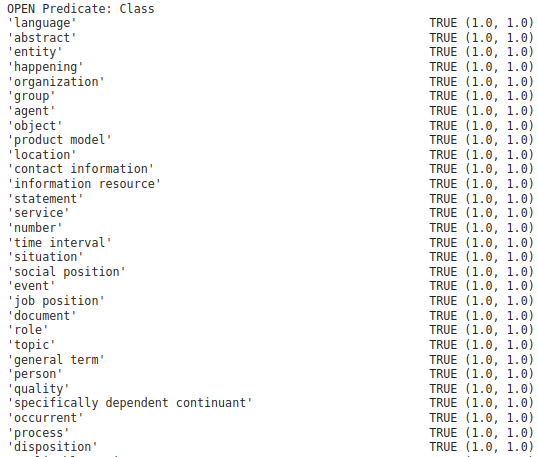}}
    \caption{Log of all Class and Instance nodes}
\end{figure*}

\begin{figure*}[!ht]
    \centering
    \noindent{\includegraphics[width=\textwidth]{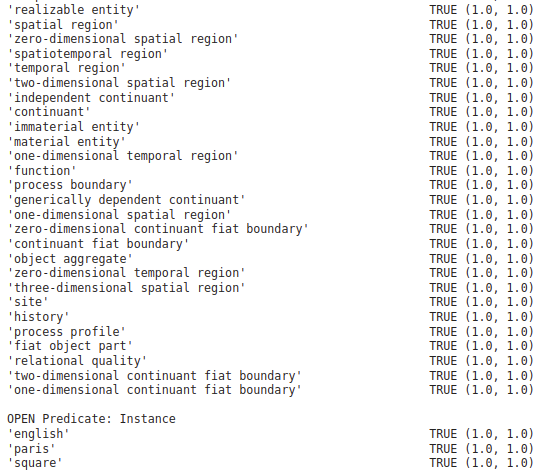}}
    \caption{Log of all Class and Instance nodes - continued}
\end{figure*}

\bibliographystyle{aaai}
\bibliography{aaai}

\end{document}